\pdfoutput=1

\documentclass[11pt]{article}

\usepackage{acl}

\usepackage{times}
\usepackage{latexsym}

\usepackage[T1]{fontenc}

\usepackage[utf8]{inputenc}

\usepackage{microtype}

\usepackage{amsmath}
\usepackage{amsfonts}

\usepackage{booktabs}
\usepackage{graphicx}
\usepackage{subcaption}
\usepackage{caption}
\usepackage{pythonhighlight}
\usepackage{wasysym}
\usepackage{algorithm}
\usepackage{algpseudocode}

\DeclareMathOperator*{\argmax}{arg\,max}

\title{Instructions for *ACL Proceedings}

\title{WeaNF: Weak Supervision with Normalizing Flows}

\author{Andreas Stephan \\
  University of Vienna \\
  \texttt{andreas.stephan@univie.ac.at$  $} \\\And
  Benjamin Roth \\
  University of Vienna \\
  \texttt{benjamin.roth@univie.ac.at} \\}

\date{}

\begin{document}
\maketitle

\begin{abstract}
A popular approach to decrease the need for costly manual annotation of large data sets is weak supervision, which introduces problems of noisy labels, coverage and bias.
Methods for overcoming these problems have either relied on discriminative models, trained with cost functions specific to weak supervision, and more recently, generative models, trying to  model the output of the automatic annotation process.
In this work, we explore a novel direction of generative modeling for weak supervision:
Instead of modeling the output of the annotation process (the labeling function matches), we generatively model the input-side data distributions (the feature space) covered by labeling functions.
Specifically,  we estimate a density for each weak labeling source, or labeling function, by using normalizing flows.
An integral part of our method is the flow-based modeling of multiple simultaneously matching labeling functions, and therefore phenomena such as labeling function overlap and correlations are captured.
We analyze the effectiveness and modeling capabilities on various commonly used weak supervision data sets, and show that weakly supervised normalizing flows compare favorably to standard weak supervision baselines. 
\end{abstract}

\section{Introduction}

Currently an important portion of research in natural language processing is devoted to the goal of reducing or getting rid of large labeled datasets.  
Recent examples include language model fine-tuning \citep{devlin2019bert}, transfer learning \citep{zoph2016transfer} or few-shot learning \citep{brown2020language}.
Another common approach is weakly supervised learning.  The idea is to make use of human intuitions or already acquired human knowledge to create weak labels.  
Examples of such sources are keyword lists, regular expressions, heuristics or independently existing curated data sources, e.g. a movie database if the task is concerned with TV shows. 
While the resulting labels are noisy, they provide a quick and easy way to create large labeled datasets. 
In the following, we use the term labeling functions, introduced in \citet{DBLP:journals/corr/abs-1711-10160}, to describe functions which create weak labels based on the notions above.

Throughout the weak supervision literature generative modeling ideas are found \citep{takamatsu-etal-2012-reducing,alfonseca-etal-2012-pattern,DBLP:journals/corr/abs-1711-10160}.
Probably the most popular example of a system using generative modeling in weak supervision is the data programming paradigm of Snorkel \citep{DBLP:journals/corr/abs-1711-10160}. 
It uses correlations within labeling functions to learn a graph capturing dependencies between labeling functions and true labels.  

However, such an approach does not directly model biases of weak supervision reflected in the feature space.  
In order to directly model the relevant aspects in the feature space of a weakly supervised dataset, we investigate the use of density estimation using normalizing flows. 
More specifically, in this work, we model probability distributions over the input space induced by \emph{labeling functions}, and combine those distributions for better weakly supervised prediction.

We propose and examine four novel models for weakly supervised learning based on normalizing flows (\textbf{WeaNF-*}):
Firstly,  we introduce a \textbf{standard} model \textbf{WeaNF-S}, where each labeling function is represented by a multivariate normal distribution,  and its \textbf{iterative} variant \textbf{WeaNF-I}.
Furthermore \textbf{WeaNF-N} additionally learns the \textbf{negative} space, i.e. a density for the space where the labeling function does not match, and a \textbf{mixed} model, \textbf{WeaNF-M}, where correlations of sets of labeling functions are represented by the normalizing flow.
As a consequence, the classification task is a two step procedure. The first step estimates the densities, and the second step aggregates them to model label prediction.  
Multiple alternatives are discussed and analyzed.

We benchmark our approach on several commonly used weak supervision datasets. 
The results highlight that our proposed generative approach is competitive with standard weak supervision methods. 
Additionally the results show that smart aggregation schemes prove beneficial.

In summary, our contributions are i) the development of multiple models based on normalizing flows for weak supervision combined with density aggregation schemes, ii) a quantitative and qualitative analysis highlighting opportunities and problems and iii) an implementation of the method\footnote{\url{https://github.com/AndSt/wea_nf}}.
To the best of our knowledge we are the first to use normalizing flows to generatively model labeling functions.

\section{Background and Related Work}
\label{sec:related_work}

We split this analysis into a weak supervision and a normalizing flow section as we build upon these two areas.

\textbf{Weak supervision. } A fundamental problem in machine learning is the need for massive amounts of manually labeled data.  Weak supervision provides a way to counter the problem. The idea is to use human knowledge to produce noisy, so called weak labels.  Typically, keywords, heuristics or knowledge from external data sources is used. The latter is called distant supervision \citep{10.5555/645634.663209,mintz-etal-2009-distant}.
In \citet{DBLP:journals/corr/abs-1711-10160}, data programming is introduced, a paradigm to create and work with weak supervision sources programmatically. 
The goal is to learn the relation between weak labels and the true unknown labels \citep{DBLP:journals/corr/abs-1711-10160,varma2019learning,DBLP:journals/corr/BachHRR17,DBLP:journals/corr/abs-1911-09860}. In \citet{2020} the authors use iterative modeling for weak supervision.
Software packages such as SPEAR \citep{abhishek2021spear}, WRENCH \citep{zhang2021wrench} and Knodle \citep{sedova2021knodle} allow a modular use and comparison of weak supervision methods.
A recent trend is to use additional information to support the learning process.  \citet{DBLP:journals/corr/abs-1911-09860} allow labeling functions to assign a score to the weak label. In \citet{ratner2018training} the human provided class balance is used.
Additionally \citet{awasthi2020learning,karamanolakis2021selftraining} use semi-supervised methods for weak supervision, where the idea is to use a small amount of labeled data to steer the learning process. 
\\
\\
\textbf{Normalizing flows.}
While the concept of normalizing flows is much older, \citet{rezende2016variational} introduced the concept to deep learning. In comparison to other generative neural networks, such as Generative Adversarial networks \citep{goodfellow2014generative} or Variational Autoencoders \citep{kingma2014autoencoding}, normalizing flows provide a tractable way to model high-dimensional distributions.  
So far, normalizing received rather little attention in the natural language processing community. Still, \citet{tran2019discrete} and \citet{ziegler2019latent} applied them successfully to language modeling.
An excellent overview over recent normalizing flow research is given in \citet{papamakarios2021normalizing}.
Normalizing flows are based on the change of variable formula, which uses a bijective function $g: Z \rightarrow X$ to transform a base distribution $Z$ into a target distribution $X$:
\begin{align*}
p_X(x) = p_Z(z) \left| \det \left( \frac{\partial g(z)}{\partial z^T} \right) \right|^{-1}
\end{align*}
where $Z$ is typically a simple distribution, e.g. multivariate normal distribution, and $X$ is a complicated data generating distribution. 
Typically, a neural network learns a function $f: X \rightarrow Z$ by minimizing the KL-divergence between the data generating distribution and the simple base distribution.  As described in \citet{papamakarios2021normalizing} this is achieved by minimizing negative log likelihood
\begin{align*}
\log p_X(x) = \log p_Z(f(x)) + \log \left| \det \left( \frac{\partial f(x)}{\partial x^T} \right) \right|
\end{align*}
The tricky part is to design efficient architectures which are invertible and provide an easy and efficient way to compute the determinant.  The composition of bijective functions is again bijective which enables deep architectures $f= f_1 \circ \dots \circ f_n$.
Recent research focuses on the creation of more expressive transformation modules \citep{lu2021implicit}. In this work, we make use of an early, but well established model, called RealNVP \cite{dinh2017density}. In each layer, the input $x$ is split in half and transformed according to
\begin{align}
y_{1:d}      &= x_{1:d} \\
y_{d+1:D} &= x_{d+1:D} \odot \exp \left(  s(x_{1:d}) \right) + t \left(x_{1:d} \right) \label{eq:real_nvp}
\end{align}
where $\odot$ is the pointwise multiplication and $s$ and $t$ neural networks. Using this formulation to realize a layer $f_i$, it is easy and efficient to compute the inverse and the determinant.

\begin{figure*}
\centering
\begin{subfigure}{.48\textwidth}
  \centering
  \includegraphics[width=\linewidth]{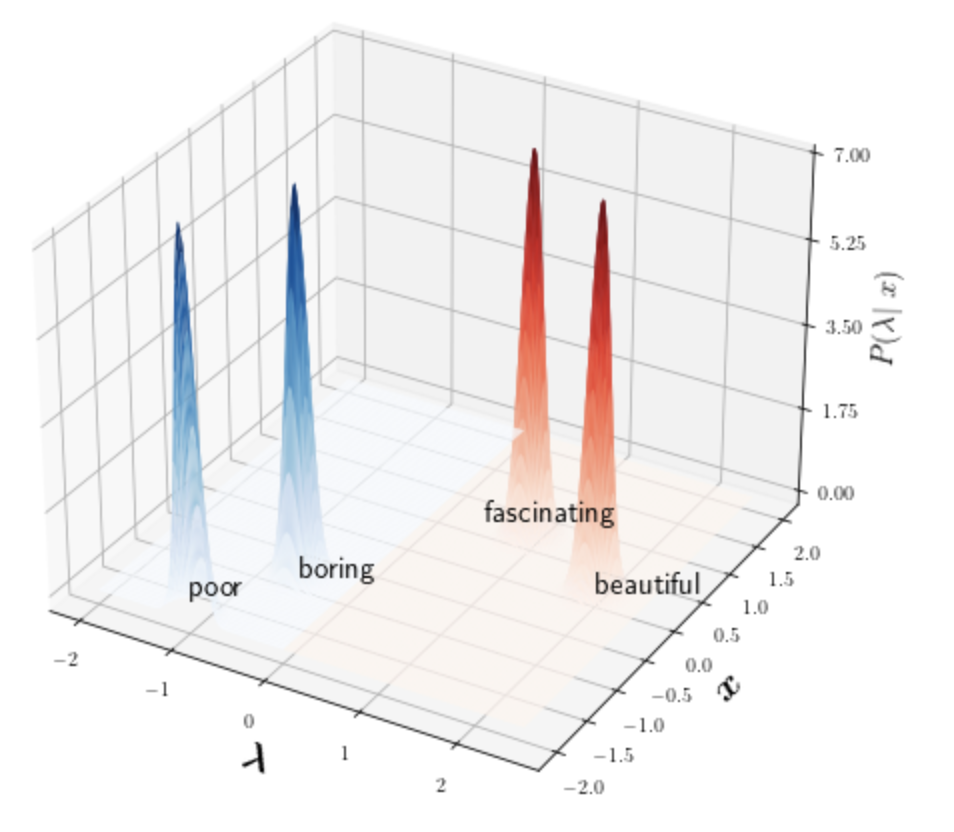}
  \caption{Schematic view of the densities estimated by \textbf{WeaNF-S/I}.  The concatenated input $[\boldsymbol{x}; \boldsymbol{\lambda} ]$ is fed into the flow to learn the probability $P(x | \lambda)$. The graph shows the posterior $P(\lambda | x)$.}
  \label{fig:sub1}
\end{subfigure}%
$ $ $ $ $ $ $ $
\begin{subfigure}{.48\textwidth}
  \centering
  \includegraphics[width=\linewidth]{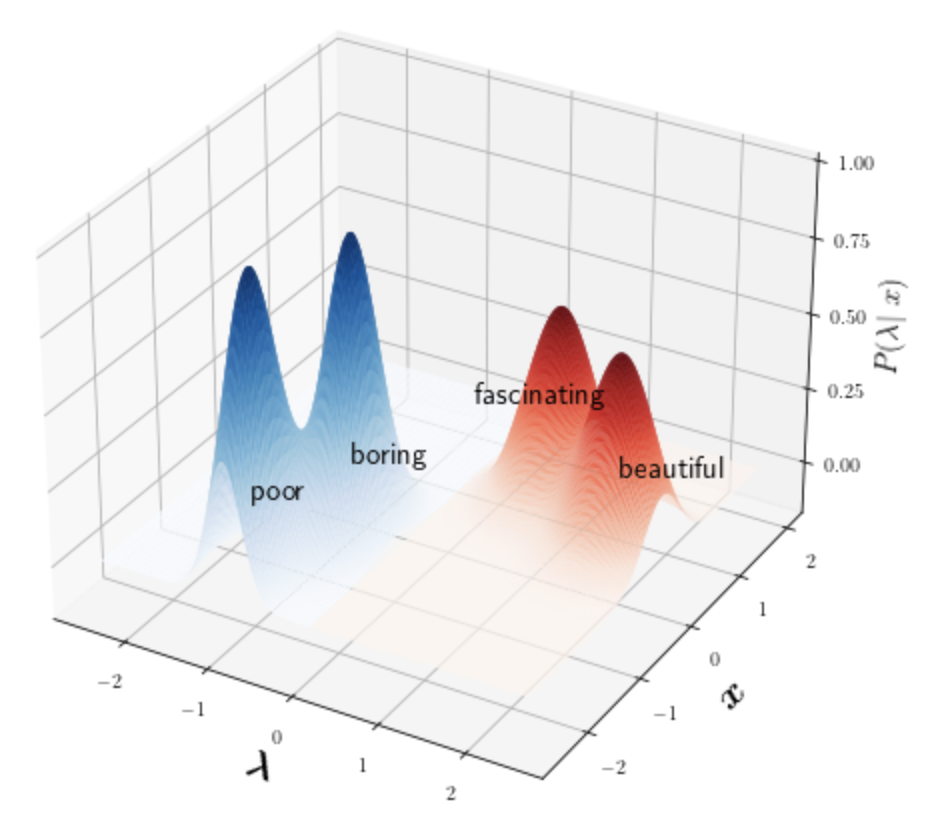}
  \caption{\textbf{WeaNF-N} and \textbf{WeaNF-M} aim to smoothen the probability space, aiming to generalize more robustly to instances not directly matched by labeling functions.}
  \label{fig:sub2}
\end{subfigure}
\caption{Schematic overview of \textbf{WeaNF-*}.  The $X-$axis represents the labeling function embedding $\boldsymbol{\lambda}$, the $Y-$axis the text input $\boldsymbol{x}$. 
The $Z-$axis represents the learned density related to a labeling function.  
In this example we use the task sentiment analysis and keyword search as labeling functions. 
Blue denotes a negative sentiment and red a positive sentiment.}
\label{fig:test}
\end{figure*}

Normalizing flows were used for semi-supervised classification \citep{izmailov2019semisupervised,atanov2020semiconditional} but not for weakly supervised learning, which we introduce in the next chapter.

\section{Model Description}
\label{sec:modelling}

In this section the models are introduced. 
The following example motivates the idea.
Consider the sentence $s$, "The movie was fascinating, even though the graphics were poor, maybe due to a low budget.", the task sentiment analysis and labeling functions given by the keywords "fascinating" and "poor". 
Furthermore, "fascinating" is associated with the class POS, and "poor" with the class NEG.
We aim to learn a neural network, which translates the complex object, text and a possible labeling function match, to a density, in the current example $P(s | \text{fascinating})$ and $P(s | \text{poor})$.
We combine this information using basic probability calculus to make a classification prediction.

Multiple models are introduced. The standard model \textit{WeaNF-S} naively learns to represent each labeling function as a multivariate normal distribution.
In order to make use of unlabeled data,  i.e. data where no labeling function matches, we iteratively apply the standard model \textit{(WeaNF-I)}.
Based on the observation that labeling functions overlap, we derive \textit{WeaNF-N} modeling the negative space, i.e. the space where the labeling function does not match and the mixed model, \textit{WeaNF-M}, using a common space for single labeling functions and the intersection of these.
Furthermore, multiple aggregation schemes are used to combine the learned labeling function densities. See table \ref{table:aggregation_overview} for an overview.
\\
\\
Before we dive into details, we introduce some notation.
From the set of all possible inputs $\mathcal{X}$, e.g. texts, we denote an input sample by $x$ and its corresponding vector representation by $\boldsymbol{x}$.
The set of $t$ labeling functions is $T=\{\lambda_1, \dots, \lambda_t \}$ and the classes are $Y=\{y_1, \dots, y_c\}$.  
Each labeling function $\lambda: \mathcal{X} \rightarrow \emptyset \cup \{y\}$ maps the input to a specific class $y \in Y$ or abstains from labeling. 
In some of our models, we also associate an embedding with each labeling function, which we denote by $\boldsymbol{\lambda} \in \mathbb{R}^h$.
The set of labeling functions corresponding to label $y$ is $T_y$.

 \begin{table*}
\centering
\begin{tabular}{lcccc}
\toprule
  & $P(y | x) \propto$ & WeaNF-S/I & WeaNF-N & WeaNF-M \\
\midrule
   Maximum & $\max_{\lambda \in T_y} P_\theta(x | \lambda)$ & $\surd$ & & $\surd$ \\
   Union & $\sum_{\lambda \in T_y} P(\lambda | x)$ & & $\surd$ & \\
   NoisyOr & $1 - \prod_{\lambda \in T_y} \left( 1 - P(\lambda | x) \right)$ & & $\surd$ & \\
   Simplex &  $P\left( \left[\boldsymbol{x}; \frac{1}{| T_y |} \sum_{\lambda \in T_y}  \boldsymbol{\lambda} \right] \right)$ & & & $\surd$ \\
\bottomrule
\end{tabular}
\caption{Overview over the used aggregation schemes. Note that $P(\lambda | x)$ is only accessible with WeaNF-N (see equation \ref{eq:neg_bayes_transition}). 
Bold symbols denote vector representations. }
\label{table:aggregation_overview}
\end{table*}

\textbf{WeaNF-S/I.} The goal of the standard model is to learn a distribution $P(x | \lambda)$ for each labeling function $\lambda$. 
Similarly to \citet{atanov2020semiconditional} in semi-supervised learning, we use a randomly initialized embedding $\boldsymbol{\lambda} \in \mathbb{R}^{h}$ to create a representation for each labeling function in the input space.  
We concatenate input and labeling function vector and provide it as input to the normalizing flow, thus learning $P([\boldsymbol{x};\boldsymbol{\lambda}_i])$, where $[\cdot ]$ describes the concatenation operation.  
A standard RealNVP \cite{dinh2017density}, as described in section \ref{sec:related_work} is used. 
See appendix \ref{sec:app_detailed_architecture} for implementational details.
In order to use the learned probabilities to perform label prediction, an aggregation scheme is needed. For the sake of simplicity, the model predicts the label corresponding to the labeling function with the highest likelihood, $y=\argmax_{y \in Y} \max_{\lambda \in T_y} P(x | \lambda)$.

Additionally, to make use of the unlabeled data, i.e. the data points where no labeling function matches, an iterative version WeaNF-I is tested.  For this, we use an EM-like \citep{Dempster77maximumlikelihood} iterative scheme where the predictions of the model trained in the previous iteration are used as labels for the unlabeled data.  The corresponding pseudo-code is found in algorithm \ref{alg:weanf-i}.
\\
\\



\begin{algorithm}
\caption{Iterative Model (WeaNF-I)}
\begin{algorithmic}
\Require  $X_l \in \mathbb{R}^{n_l \times d}$, corresponding matches $\lambda_l \in \{0,1\}^{n_l \times t} $, 
unmatched $X_u \in \mathbb{R}^{n_u \times d}$

\State $F$ = train\_flow$(X_l, \lambda_l)$
\For{$i=1, \dots, r$}
      \State $\left(\lambda_u \right)_i = \argmax_{\lambda } F((X_u)_i; \lambda)$
      \State $X=\text{concat}(X_l, X_u)$, $\lambda=\text{concat}(\lambda_l, \lambda_u)$
      \State$ F =\text{train\_flow}(X, \lambda)$
\EndFor
\end{algorithmic}
\label{alg:weanf-i}
\end{algorithm}

\textbf{Negative Model. } In typical classification scenarios it is enough to learn $P(x | y)$ to compute a posterior $P(y | x)$ by applying Bayes' formula twice, resulting in
\begin{align}
P(y | x) = \frac{P(x | y) P(y)}{P(x | y) P(y) + P(x |\neg y) P(\neg y)} \label{eq:bayes_classification}
\end{align}
where the class prior $P(y)$  is typically approximated on the training data or passed as a parameter.
This is not possible in the current setting as often two labeling functions match simultaneously. 
In order to learn $P(\lambda | x)$, we explore a novel variant that additionally learns $P(x | \neg \lambda)$.
The learning process is similar to $P(x | \lambda)$, so a second embedding $\tilde{\boldsymbol{\lambda}}$ is introduced to represent $\neg \lambda$.  
We optimize $P([\boldsymbol{x};\boldsymbol{\lambda}]$ and $P\left(\left[\boldsymbol{x};\tilde{\boldsymbol{\lambda}} \right]\right)$ simultaneously. 
In each batch $I$, the positive sample pairs $(x_i, \lambda_i)_{i \in I}$ and negative pairs $(x_i, \lambda_j)$, sampled such that $(x_i, \lambda_j) \notin \{ (x_i, \lambda_i) \}_{i \in I}$, are used to train the network.
The number of negative samples per positive sample is an additional hyperparameter.  Now Bayes' formula can be used as in equation \ref{eq:bayes_classification} to obtain
\begin{align}
P(\lambda | x) = \frac{P(x | \lambda) P(\lambda)}{P(x | \lambda) P(y) + P(x |\neg \lambda) P(\neg \lambda)}. \label{eq:neg_bayes_transition}
\end{align}
The access to the posterior probability $P(\lambda | x)$ provides additional opportunities to model $P(y | x)$.  
After initial experimentation we settled on two options.  
A simple addition of probabilities neglecting intersection probability, equation \ref{eq:aggr_union},  which we call Union, and the NoisyOr formula, equation \ref{eq:aggr_noisyor},  which has previously shown to be effective in weakly supervised learning \cite{keith-etal-2017-identifying}:

\begin{align}
P(y|x) &\propto \sum_{\lambda \in T_y} P(\lambda | x) \label{eq:aggr_union} \\
P(y|x) &= P(\{\vee_{\lambda \in T_y} \lambda \}|x)  \\
&= 1 - \prod_{\lambda \in T_y} \left( 1 - P(\lambda | x) \right) \label{eq:aggr_noisyor}
\end{align}


\begin{table*}
\centering
\begin{tabular}{lccccc}
\toprule
 Dataset & \#Classes & \#Train / \#Test samples &  \#LF's &  Coverage(\%) & Class Balance \\
\midrule
  IMDb &       2 &         39741 / 4993 &     20 &            0.60 &       1:1 \\
 Spouse &       2 &          8530 / 1187 &      9 &            0.30 &       1:5 \\
 YouTube &       2 &           1440 / 229 &     10 &            1.66 &       1:1 \\
 SMS &       2 &           4208 / 494 &     73 &            0.51 &       1:6 \\
 Trec &       6 &           4903 / 500 &     68 &            1.73 &       1:13:14:14:9:10  \\
\bottomrule
\end{tabular}
\caption{Some basic statistics describing the datasets.  Coverage is computed on the train set by  \#matches $/$ \#samples. }
\label{table:dataset_stats}
\end{table*}

\textbf{Mixed Model.} 
It was already mentioned that it is common that two or multiple labeling functions hit simultaneously. While WeaNF-N provides access to a posterior distribution which allows to model these interactions, the goal of the mixed model WeaNF-M is to model these intersections explicitly already in the density of the normalizing flow. More specifically, we aim to learn $P(x | \{\lambda_i\}_{i \in I})$ for arbitrary index families $I$. 
Once again, the embeddings space is used to achieve this goal.
For a given sample $x$ and a family $I$ of matching labeling functions,  we uniformly sample from the simplex of all possible combinations and obtain $\boldsymbol{\lambda}_I = \sum_{i \in I} \alpha_i \boldsymbol{\lambda}_i, \alpha_i \geq 0, \sum_{i \in I} \alpha_i = 1$. 
Afterwards we concatenate the weighted sum of the labeling function embeddings $\boldsymbol{\lambda}_I$ with the input $x$ and learn $P([\boldsymbol{x}; \boldsymbol{\lambda}_I])$. 
Now that the density is able to access the intersections of labeling functions, we derive a new direct aggregation scheme.
By $\sigma_y$ we denote the simplex generated by the set of boundary points $\{\boldsymbol{\lambda}\}_{\lambda \in T_y}$.  It is important to think about this simplex, as it theoretically describes the input space where the model learns the density related to class $y$.
We use the naive but efficient variant which just computes the center of the simplex:
\begin{align}
P(y | x) & \propto P\left( \left[\boldsymbol{x}; \frac{1}{| T_y |} \sum_{\lambda \in T_y}  \boldsymbol{\lambda} \right] \right)
\label{eq:approx_aggr_integral}
\end{align}


\textbf{Implementation. } 
In practice, sampling of data points has to be handled on multiple occasions.
Empirically and during the inspection of related implementations, e.g. the Github repository accompanying \citet{atanov2020semiconditional}, we found that it is beneficial if every labeling function is seen equally often during training. It supports preventing a biased density towards specific labeling functions. 
When training WeaNF-N, the negative space is much larger than the actual space, so an additional hyperparameter controlling the amount of negative samples is needed.
WeaNF-M aims to model intersecting probabilities directly. Most intersections occur too rarely to model a reasonable density.  Thus we decided to only take co-occures into account which occur more often than a certain threshold. See appendix \ref{sec:app_lf_correlation} to get a feeling for the correlations in the used datasets.

\begin{table*}
\centering
\begin{tabular}{lccccc}
\toprule
& IMDb & Spouse$(F_1)$ & YouTube & SMS $(F_1)$ & Trec \\
\midrule
MV & 56.84 &   49.87 & 81.66 & 56.1 &  61.2 \\
MV + MLP & 73.20 &   29.96 & \textbf{92.58} & 92.41 & 53.27 \\
DP + MLP & 67.79 &   57.05 & 88.79 & 84.40 & 43.00 \\
\hline
  WeaNF-S &  73.06 &   52.28 &  89.08 &  86.71 &  67.4 \\
  WeaNF-I &  \textbf{74.08} &   \textbf{57.96} &  89.08 &  \textbf{93.54} &  \textbf{67.8} \\
  WeaNF-N (NoisyOr)& 72.96 &   54.60 & 90.83 & 79.63 &  54.8 \\
  WeaNF-N (Union) & 71.98 &   50.83 & 91.70 & 83.48 &  60.2 \\
  WeaNF-M (Max) & 70.16 &   55.16 & 85.15 & 88.23 &  49.8 \\
  WeaNF-M (Simplex) & 63.53 &   56.91 & 86.03 & 76.29 &  25.4\\
\bottomrule
\end{tabular}
\caption{Comparison of baselines to our model variants.  The numbers reflect accuracies, or $F_1$-scores, where explicitly mentioned. Names in parenthesis describe the aggregation mechanism.}
\label{table:main_exp}
\end{table*}

\section{Experiments}

In order to analyze the proposed models experiments on multiple standard weakly supervised classification problems are performed.  
In the following, we introduce datasets, baselines and training details.

\subsection{Datasets}

Within our experiments, we use five classification tasks. Table \ref{table:dataset_stats} gives an overview over some key statistics.  Note that these might differ slightly compared to other papers due to the removal of duplicates.  For a more detailed overview of our preprocessing steps, see appendix \ref{subsec:app_preprocessing}.

The first dataset is \textbf{IMDb} (Internet Movie Database) and the accompanying sentiment analysis task \cite{maas-etal-2011-learning}. 
The goal is to classify whether a movie review describes a positive or a negative sentiment. 
We use $10$ positive and $10$ negative keywords as labeling functions.  
See Appendix \ref{subsec:app_imdb_keywords} for a detailed description.

The second dataset is the \textbf{Spouse} dataset \cite{Corney2016WhatDA}. 
The task is to classify whether a text holds a spouse relation, e.g. "Mary is married to Tom".  
Here, $90\%$ of the samples belong to the no-relation class, so we use macro-$F_1$ score to evaluate the performance.
As the third dataset another binary classification problem is given by the  \textbf{YouTube} Spam \cite{7424299} dataset.
The model has to decide whether a YouTube comment is spam or not. For both, the Spouse and the YouTube dataset, the labeling functions are provided by the Snorkel framework \cite{DBLP:journals/corr/abs-1711-10160}.

The \textbf{SMS} Spam detection dataset \cite{Almeida2011ContributionsTT}, we abbreviate by SMS, also asks for spam but in the private messaging domain.  The dataset is quite skewed, so once again macro-$F_1$ score is used.
Lastly, a multi-class dataset, namely \textbf{TREC-6} \cite{li-roth-2002-learning}, is used. The task is to classify questions into six categories, namely Abbreviation, Entity, Description, Human and Location. 
The labeling functions provided by \cite{awasthi2020learning} are used for the SMS and the TREC dataset.
We took the preprocessed versions of the data available within the Knodle weak supervision programming framework \cite{sedova2021knodle}.  

\subsection{Baselines}
Three baselines are used.
While there are many weak supervision systems, most use additional knowledge to improve performance. 
Examples are class balance \cite{DBLP:journals/corr/abs-1911-09860}, semi-supervised learning with very little labels \cite{awasthi2020learning,karamanolakis2021selftraining} or multi-task learning \cite{ratner2018training}. 
To ensure a fair comparison,  only baselines are used that solely take input data and labeling function matches into account.
First we use majority voting (MV) which takes the label where the most rules match.  For instances where multiple classes have an equal vote or where no labeling function matches, a random vote is taken. Secondly, a multi-layer perceptron (MLP) is trained on top of the labels provided by majority vote.
The third baseline uses the data programming (DP) paradigm. 
More explicitly, we use the model introduced by \citet{ratner2018training} implemented in the Snorkel \cite{DBLP:journals/corr/abs-1711-10160} programming framework.  
It performs a two-step approach to learning. 
Firstly, a generative model is trained to learn the most likely correlation between labeling functions and unknown true labels.  
Secondly, a discriminative model uses the labels of the generative model to train a final model. The same MLP as for second baseline is used for the final model.

\subsection{Training Details}

Text input embeddings are created with the SentenceTransformers library \cite{reimers-2019-sentence-bert} using the \textit{bert-base-nli-mean-tokens} model.
They serve as input to the baselines and the normalizing flows.
Hyperparameter search is performed via grid search over learning rates of $\{1e-5, 1e-4\}$, weight decay of $\{1e-2, 1e-3\}$ and epochs in $\{30,50,100,300,450\}$, and label embedding dimension in $10,15,20$ times the number of classes. 
Additionally, the number of layers is in $\{6, 8\}$, and the negative sampling value for WeaNF is in $\{2, 3\}$.
The full set up ran $30$ hours on a single GPU on a DGX $1$ server.

\section{Analysis}

The analysis is divided into three parts. Firstly,  a general discussion of the results is given. Secondly, an analysis of the densities predicted by WeaNF-N is shown and lastly, a qualitative analysis is performed.

\begin{table*}
\centering
\fontsize{7.5}{8}\selectfont
\begin{tabular}{lp{45mm}lllrr}
\toprule
 Labeling Function & Example & Dataset & $P(x | \lambda)$ & Label $(\lambda)$ & Gold &  Prediction \\
\midrule
won .* claim & ...won ... call ...& SMS & $\uparrow$ & Spam & Spam & Spam \\
.* I'll .* & sorry, \textbf{I'll} call later	& SMS & $\uparrow$  & No Spam & No Spam& No Spam \\
.* i .* & \textbf{i} just saw ron burgundy captaining a party boat so yeah & SMS & $\downarrow$ & No Spam & No Spam& No Spam \\ 

(explain|what) .* mean .* & \textbf{What} does the abbreviation SOS \textbf{mean} ? & Trec & $\uparrow$ & DESCR & ABBR & DESCR \\
(explain|what) .* mean .* & What are Quaaludes ?	&Trec & $\uparrow$ &  DESCR & DESCR & DESCR \\
who.* & \textbf{Who} was the first man to ... Pacific Ocean ?	& Trec & $\downarrow$& HUMAN & HUMAN & HUMAN \\

check .* out .*& \textbf{Check out} this video on YouTube:  & YouTube & $\uparrow$ & Spam & Spam & Spam \\
\#words < 5& subscribe my  & YouTube & $\uparrow$ & Spam & Spam & No Spam \\
.* song .* & This \textbf{Song} will never get old	 & YouTube& $\downarrow$ & No Spam & No Spam &  No Spam \\ 

.* dreadful .* & ...horrible performance .... annoying & IMDb & $\uparrow$ & NEG & NEG & NEG\\ 
.* hilarious .* & ...liked the movie...funny catchphrase...WORST...low grade... & IMDb & $\uparrow$ & POS & NEG & POS \\
.* disappointing .* & don't understand stereotype ... goofy .. & IMDb & $\downarrow$ & NEG & NEG & POS \\  

.* (husband|wife) .* & ...Jill.. she and her \textbf{husband}...  & Spouse & $\uparrow$ & Spouses & Spouses & Spouses \\ 
.* married .* & ... asked me to marry him and I said yes! & Spouse & $\uparrow$ & Spouses & No Spouses & Spouses \\ 
family word & Clearly excited, Coleen said: 'It's my eldest \textbf{son }Shane and Emma. & Spouse & $\downarrow$ & No Spouses & No Spouses & No Spouses \\ 

\bottomrule
\end{tabular}

\caption{
Examples selected from the $10$ most likely ($\uparrow$) and $10$ most unlikely ($\downarrow$) combinations of sentences and labeling functions, using the density $P(x| \lambda)$ provided by WeaNF-I. 
Labeling function matches are bold. We observe that the flow often generalizes to unmatched examples.
We slightly simplified some rules and shortened some texts in order to fit the page size.  
}
\label{table:examples}
\end{table*}

\subsection{Overall Findings}

Table \ref{table:main_exp} exposes the main evaluation.  The horizontal line separates the baselines from our models. 
For WeaNF-N and WeaNF-M, no iterative schemes were trained. 
This enables a direct comparison to the standard model WeaNF-I.

Interestingly, the combination of Snorkel and MLP's is often not performing competitively.
In the IMDb data set there is barely any correlation between labeling functions, complicating Snorkel's approach.  
The large number of labeling functions e.g. Trec, SMS,  could also complicate correlation based approaches. Appendix \ref{sec:app_lf_correlation} shows correlation graphs.

As indicated by the bold numbers, the WeaNF-I is the best performing model. 
Only on the YouTube dataset, an iterative scheme could not improve the results. 
Related to this observation, in \citet{2020} the authors achieve promising results using iterative discriminative modeling for semi-supervised weak supervision.


WeaNF-N outperforms the standard model in three out of five datasets. We observe that these are the datasets with a large amount of labeling functions. Possibly, this biases the model towards a high value of $P(x | \neg \lambda)$ which confuses the prediction.

The simplex aggregation scheme only outperforms the maximum aggregation on two out of five datasets. 
We infer that the probability density over the labeling function input space is not smooth enough. 
Ideally, the simplex method should always have a high confidence in the prediction of a labeling function $\lambda$ if its confident on the non-mixed embedding $\boldsymbol{\lambda}$ which is what Max is doing.





\begin{table}
\small
\centering
\begin{tabular}{llllll}
\toprule
& \textbf{IMDb}&\textbf{Spouse}&\textbf{YouTube} & \textbf{SMS }& \textbf{Trec} \\
\midrule
Acc      &  72.38 &  74.04 &  78.17 &  88.71 &  72.63 \\
$P$ &   5.93 &    5.1 &  38.95 &   23.3 &  13.65 \\
$R$   &  37.53 &  39.31 &  55.01 &  44.34 &  61.07 \\
$F_1$        &  10.25 &   9.02 &  45.61 &  30.55 &  22.31 \\
Cov       &   4.31 &   5.74 &  19.31 &   3.01 &   4.39 \\
\bottomrule
\end{tabular}
\caption{Evaluation of the labeling function prediction $P(\lambda | x)$. Precision, Recall and $F_1$ score are computed via the weighted average of the statistics of all labeling functions.  Coverage is computed as \#matches$/$\#all possible matches.}
\label{table:lf_accs}
\end{table}

\begin{table}
\footnotesize
\begin{tabular}{lllll}
\toprule
Dataset & Labeling Fct. & Cov(\%) & Prec & Recall \\ 
\midrule
IMDb & *boring* & 5.8 & 13.12 & 26.87 \\ 
Spouse & family word  & 9. 0 & 16.53 & 35.96  \\ 
YouTube & *song* & 23.58 & 56.72 & 70.73   \\ 
SMS & won *call*  & 0.81 & 66.67 & 1.0  \\ 
Trec & how.*much &  2.4 & 60.0 & 75.0  \\ 
\bottomrule
\end{tabular}
\caption{Statistics for the labeling functions obtaining the highest $F_1$ score for the prediction $P(\lambda | x)$, using the WeaNF (NoisyOr) model.}
\label{table:best_lf}
\end{table}

\begin{table}
\footnotesize
\begin{tabular}{lllll}
\toprule
Dataset & Labeling Fct.  & Cov(\%) & Prec & Recall \\ 
\midrule
IMDb & *imaginative* & 0.42 & 0.77 & 52.38 \\ 
Spouse & spouse keyword & 14.5 & 0 & 0 \\ 
YouTube & person entity & 2.62 & 6.45 & 33.33 \\ 
SMS & I .* miss & 0.6 & 0 & 0 \\ 
Trec & what is .* name & 2.2& 2.26 & 100 \\ 
\bottomrule
\end{tabular}
\caption{Same as table \ref{table:best_lf}, but here the labeling functions obtaining the lowest $F_1$ score are shown. Only those are taken into account which occur more often than $10$ times in the test set. }
\label{table:worst_lf}
\end{table}

\subsection{Density Analysis}

We divide into a global analysis and a local, i.e. a per-labeling function, analysis.  
Table \ref{table:lf_accs} provides some global statistics, table \ref{table:best_lf} and \ref{table:worst_lf} subsequently show statistics related to the best and worst performing labeling function estimations.  
In the local analysis a labeling function is predicted if $P(\lambda | x) \geq 0.5$.
The WeaNF-N model is used because it is the only model with direct access to $P(\lambda | x)$.  

It is important to mention that in the local analysis, a perfect prediction of the matching labeling function is not wanted, as this would mean that there is no generalization. 
Thus, a low precision might be necessary for generalization, and a the recall would indicate how much of the original semantic or syntactic meaning of a labeling function is retained.

Interestingly,  while the overall performance of WeaNF-N is competitive on the IMDb and the Spouse data sets, it is failing to predict the correct labeling function. 
One explanation might be that these are the data sets where the texts are substantially longer which might be complicated to model for normalizing flows. 
In table \ref{table:worst_lf} typically the worst performing approximation of labeling function matches seems to be due to low coverage.  An exception is the the Spouse labeling function.

\subsection{Qualitative Analysis}

In table \ref{table:examples} a number of examples are shown. We manually inspected samples with a very high or low density value. 
Note that density values related to $P(x | \lambda), \lambda \in T_y$ are functions $f$ taking arbitrary values which only have to satisfy $\mathbb{E}_{x:\lambda(x)=y}[f(x)]=1$.

We observed the phenomenon that either the same labeling functions take the highest density values $P(x | \lambda)$ or that a single sample often has a high likelihood for multiple labeling functions.
In the table \ref{table:examples} one can find examples where the learned flows were able to generalize from the original labeling functions. For example, for the IMDb dataset, it detects the meaning "funny" even though the exact keyword is "hilarious".

\section{Conclusion}

This work explores the novel use of normalizing flows for weak supervision.  
The approach is divided into two logical steps. 
In the first step,  normalizing flows are employed to learn a probability distribution over the input space related to a labeling function. 
Secondly, principles from basic probability calculus are used to aggregate the learned densities and make them usable for classification tasks.
Motivated by aspects of weakly supervised learning, such as labeling function overlap or coverage, multiple models are derived each of which uses the information present in the latent space differently.
We show competitive results on five weakly supervised classification tasks.
Our analysis shows that the flow-based representations of labeling functions successfully generalize to samples otherwise not covered by labeling functions.

\section*{Acknowledgements}

This research was funded by the WWTF through the project ”Knowledge-infused Deep Learning for Natural Language Processing” (WWTF Vienna Research Group VRG19-008), and by the Deutsche Forschungsgemeinschaft (DFG, German Research Foundation) - RO 5127/2-1.

\bibliography{anthology,custom}

\begin{thebibliography}{36}
\expandafter\ifx\csname natexlab\endcsname\relax\def\natexlab#1{#1}\fi

\bibitem[{Abhishek et~al.(2021)Abhishek, Ingole, Laturia, Dorna, Maheshwari,
  Ramakrishnan, and Iyer}]{abhishek2021spear}
Guttu~Sai Abhishek, Harshad Ingole, Parth Laturia, Vineeth Dorna, Ayush
  Maheshwari, Ganesh Ramakrishnan, and Rishabh Iyer. 2021.
\newblock \href {http://arxiv.org/abs/2108.00373} {{SPEAR : Semi-supervised
  Data Programming in Python}}.

\bibitem[{Alberto et~al.(2015)Alberto, Lochter, and Almeida}]{7424299}
T~C Alberto, J~V Lochter, and T~A Almeida. 2015.
\newblock \href {https://doi.org/10.1109/ICMLA.2015.37} {{TubeSpam: Comment
  Spam Filtering on YouTube}}.
\newblock In \emph{2015 IEEE 14th International Conference on Machine Learning
  and Applications (ICMLA)}, pages 138--143.

\bibitem[{Alfonseca et~al.(2012)Alfonseca, Filippova, Delort, and
  Garrido}]{alfonseca-etal-2012-pattern}
Enrique Alfonseca, Katja Filippova, Jean-Yves Delort, and Guillermo Garrido.
  2012.
\newblock \href {https://aclanthology.org/P12-2011} {{Pattern Learning for
  Relation Extraction with a Hierarchical Topic Model}}.
\newblock In \emph{Proceedings of the 50th Annual Meeting of the Association
  for Computational Linguistics (Volume 2: Short Papers)}, pages 54--59, Jeju
  Island, Korea. Association for Computational Linguistics.

\bibitem[{Almeida et~al.(2011)Almeida, Hidalgo, and
  Yamakami}]{Almeida2011ContributionsTT}
Tiago~A Almeida, J~M~G Hidalgo, and A~Yamakami. 2011.
\newblock {Contributions to the study of SMS spam filtering: new collection and
  results}.
\newblock In \emph{DocEng '11}.

\bibitem[{Atanov et~al.(2020)Atanov, Volokhova, Ashukha, Sosnovik, and
  Vetrov}]{atanov2020semiconditional}
Andrei Atanov, Alexandra Volokhova, Arsenii Ashukha, Ivan Sosnovik, and Dmitry
  Vetrov. 2020.
\newblock \href {http://arxiv.org/abs/1905.00505} {{Semi-Conditional
  Normalizing Flows for Semi-Supervised Learning}}.

\bibitem[{Awasthi et~al.(2020)Awasthi, Ghosh, Goyal, and
  Sarawagi}]{awasthi2020learning}
Abhijeet Awasthi, Sabyasachi Ghosh, Rasna Goyal, and Sunita Sarawagi. 2020.
\newblock \href {http://arxiv.org/abs/2004.06025} {{Learning from Rules
  Generalizing Labeled Exemplars}}.

\bibitem[{Bach et~al.(2017)Bach, He, Ratner, and
  R{\'{e}}}]{DBLP:journals/corr/BachHRR17}
Stephen~H Bach, Bryan~Dawei He, Alexander Ratner, and Christopher R{\'{e}}.
  2017.
\newblock \href {http://arxiv.org/abs/1703.00854} {{Learning the Structure of
  Generative Models without Labeled Data}}.
\newblock \emph{CoRR}, abs/1703.0.

\bibitem[{Brown et~al.(2020)Brown, Mann, Ryder, Subbiah, Kaplan, Dhariwal,
  Neelakantan, Shyam, Sastry, Askell, Agarwal, Herbert-Voss, Krueger, Henighan,
  Child, Ramesh, Ziegler, Wu, Winter, Hesse, Chen, Sigler, Litwin, Gray, Chess,
  Clark, Berner, McCandlish, Radford, Sutskever, and
  Amodei}]{brown2020language}
Tom~B Brown, Benjamin Mann, Nick Ryder, Melanie Subbiah, Jared Kaplan, Prafulla
  Dhariwal, Arvind Neelakantan, Pranav Shyam, Girish Sastry, Amanda Askell,
  Sandhini Agarwal, Ariel Herbert-Voss, Gretchen Krueger, Tom Henighan, Rewon
  Child, Aditya Ramesh, Daniel~M Ziegler, Jeffrey Wu, Clemens Winter,
  Christopher Hesse, Mark Chen, Eric Sigler, Mateusz Litwin, Scott Gray,
  Benjamin Chess, Jack Clark, Christopher Berner, Sam McCandlish, Alec Radford,
  Ilya Sutskever, and Dario Amodei. 2020.
\newblock \href {http://arxiv.org/abs/2005.14165} {{Language Models are
  Few-Shot Learners}}.

\bibitem[{Chatterjee et~al.(2019)Chatterjee, Ramakrishnan, and
  Sarawagi}]{DBLP:journals/corr/abs-1911-09860}
Oishik Chatterjee, Ganesh Ramakrishnan, and Sunita Sarawagi. 2019.
\newblock \href {http://arxiv.org/abs/1911.09860} {{Data Programming using
  Continuous and Quality-Guided Labeling Functions}}.
\newblock \emph{CoRR}, abs/1911.0.

\bibitem[{Corney et~al.(2016)Corney, Albakour, Martinez-Alvarez, and
  Moussa}]{Corney2016WhatDA}
D~Corney, M-Dyaa Albakour, Miguel Martinez-Alvarez, and Samir Moussa. 2016.
\newblock {What do a Million News Articles Look like?}
\newblock In \emph{NewsIR@ECIR}.

\bibitem[{Craven and Kumlien(1999)}]{10.5555/645634.663209}
Mark Craven and Johan Kumlien. 1999.
\newblock {Constructing Biological Knowledge Bases by Extracting Information
  from Text Sources}.
\newblock In \emph{Proceedings of the Seventh International Conference on
  Intelligent Systems for Molecular Biology}, pages 77--86. AAAI Press.

\bibitem[{Dempster et~al.(1977)Dempster, Laird, and
  Rubin}]{Dempster77maximumlikelihood}
A~P Dempster, N~M Laird, and D~B Rubin. 1977.
\newblock {Maximum likelihood from incomplete data via the EM algorithm}.
\newblock \emph{JOURNAL OF THE ROYAL STATISTICAL SOCIETY, SERIES B},
  39(1):1--38.

\bibitem[{Devlin et~al.(2019)Devlin, Chang, Lee, and
  Toutanova}]{devlin2019bert}
Jacob Devlin, Ming-Wei Chang, Kenton Lee, and Kristina Toutanova. 2019.
\newblock \href {http://arxiv.org/abs/1810.04805} {{BERT: Pre-training of Deep
  Bidirectional Transformers for Language Understanding}}.

\bibitem[{Dinh et~al.(2017)Dinh, Sohl-Dickstein, and Bengio}]{dinh2017density}
Laurent Dinh, Jascha Sohl-Dickstein, and Samy Bengio. 2017.
\newblock \href {http://arxiv.org/abs/1605.08803} {{Density estimation using
  Real NVP}}.

\bibitem[{Goodfellow et~al.(2014)Goodfellow, Pouget-Abadie, Mirza, Xu,
  Warde-Farley, Ozair, Courville, and Bengio}]{goodfellow2014generative}
Ian~J Goodfellow, Jean Pouget-Abadie, Mehdi Mirza, Bing Xu, David Warde-Farley,
  Sherjil Ozair, Aaron Courville, and Yoshua Bengio. 2014.
\newblock \href {http://arxiv.org/abs/1406.2661} {{Generative Adversarial
  Networks}}.

\bibitem[{Izmailov et~al.(2019)Izmailov, Kirichenko, Finzi, and
  Wilson}]{izmailov2019semisupervised}
Pavel Izmailov, Polina Kirichenko, Marc Finzi, and Andrew~Gordon Wilson. 2019.
\newblock \href {http://arxiv.org/abs/1912.13025} {{Semi-Supervised Learning
  with Normalizing Flows}}.

\bibitem[{Karamanolakis et~al.(2021)Karamanolakis, Mukherjee, Zheng, and
  Awadallah}]{karamanolakis2021selftraining}
Giannis Karamanolakis, Subhabrata Mukherjee, Guoqing Zheng, and Ahmed~Hassan
  Awadallah. 2021.
\newblock \href {http://arxiv.org/abs/2104.05514} {{Self-Training with Weak
  Supervision}}.

\bibitem[{Keith et~al.(2017)Keith, Handler, Pinkham, Magliozzi, McDuffie, and
  O'Connor}]{keith-etal-2017-identifying}
Katherine Keith, Abram Handler, Michael Pinkham, Cara Magliozzi, Joshua
  McDuffie, and Brendan O'Connor. 2017.
\newblock \href {https://doi.org/10.18653/v1/D17-1163} {{Identifying civilians
  killed by police with distantly supervised entity-event extraction}}.
\newblock In \emph{Proceedings of the 2017 Conference on Empirical Methods in
  Natural Language Processing}, pages 1547--1557, Copenhagen, Denmark.
  Association for Computational Linguistics.

\bibitem[{Kingma and Welling(2014)}]{kingma2014autoencoding}
Diederik~P Kingma and Max Welling. 2014.
\newblock \href {http://arxiv.org/abs/1312.6114} {{Auto-Encoding Variational
  Bayes}}.

\bibitem[{Li and Roth(2002)}]{li-roth-2002-learning}
Xin Li and Dan Roth. 2002.
\newblock \href {https://aclanthology.org/C02-1150} {{Learning Question
  Classifiers}}.
\newblock In \emph{{COLING} 2002: The 19th International Conference on
  Computational Linguistics}.

\bibitem[{Lu et~al.(2021)Lu, Chen, Li, Wang, and Zhu}]{lu2021implicit}
Cheng Lu, Jianfei Chen, Chongxuan Li, Qiuhao Wang, and Jun Zhu. 2021.
\newblock \href {https://openreview.net/forum?id=8PS8m9oYtNy} {{Implicit
  Normalizing Flows}}.
\newblock In \emph{International Conference on Learning Representations}.

\bibitem[{Maas et~al.(2011)Maas, Daly, Pham, Huang, Ng, and
  Potts}]{maas-etal-2011-learning}
Andrew~L Maas, Raymond~E Daly, Peter~T Pham, Dan Huang, Andrew~Y Ng, and
  Christopher Potts. 2011.
\newblock \href {https://aclanthology.org/P11-1015} {{Learning Word Vectors for
  Sentiment Analysis}}.
\newblock In \emph{Proceedings of the 49th Annual Meeting of the Association
  for Computational Linguistics: Human Language Technologies}, pages 142--150,
  Portland, Oregon, USA. Association for Computational Linguistics.

\bibitem[{Mintz et~al.(2009)Mintz, Bills, Snow, and
  Jurafsky}]{mintz-etal-2009-distant}
Mike Mintz, Steven Bills, Rion Snow, and Daniel Jurafsky. 2009.
\newblock \href {https://aclanthology.org/P09-1113} {{Distant supervision for
  relation extraction without labeled data}}.
\newblock In \emph{Proceedings of the Joint Conference of the 47th Annual
  Meeting of the {ACL} and the 4th International Joint Conference on Natural
  Language Processing of the {AFNLP}}, pages 1003--1011, Suntec, Singapore.
  Association for Computational Linguistics.

\bibitem[{Papamakarios et~al.(2021)Papamakarios, Nalisnick, Rezende, Mohamed,
  and Lakshminarayanan}]{papamakarios2021normalizing}
George Papamakarios, Eric Nalisnick, Danilo~Jimenez Rezende, Shakir Mohamed,
  and Balaji Lakshminarayanan. 2021.
\newblock \href {http://arxiv.org/abs/1912.02762} {{Normalizing Flows for
  Probabilistic Modeling and Inference}}.

\bibitem[{Ratner et~al.(2017)Ratner, Bach, Ehrenberg, Fries, Wu, and
  R{\'{e}}}]{DBLP:journals/corr/abs-1711-10160}
Alexander Ratner, Stephen~H Bach, Henry~R Ehrenberg, Jason~Alan Fries, Sen Wu,
  and Christopher R{\'{e}}. 2017.
\newblock \href {http://arxiv.org/abs/1711.10160} {{Snorkel: Rapid Training
  Data Creation with Weak Supervision}}.
\newblock \emph{CoRR}, abs/1711.1.

\bibitem[{Ratner et~al.(2018)Ratner, Hancock, Dunnmon, Sala, Pandey, and
  R{\'{e}}}]{ratner2018training}
Alexander Ratner, Braden Hancock, Jared Dunnmon, Frederic Sala, Shreyash
  Pandey, and Christopher R{\'{e}}. 2018.
\newblock \href {http://arxiv.org/abs/1810.02840} {{Training Complex Models
  with Multi-Task Weak Supervision}}.

\bibitem[{Reimers and Gurevych(2019)}]{reimers-2019-sentence-bert}
Nils Reimers and Iryna Gurevych. 2019.
\newblock \href {https://arxiv.org/abs/1908.10084} {{Sentence-BERT: Sentence
  Embeddings using Siamese BERT-Networks}}.
\newblock In \emph{Proceedings of the 2019 Conference on Empirical Methods in
  Natural Language Processing}. Association for Computational Linguistics.

\bibitem[{Ren et~al.(2020)Ren, Li, Su, Kartchner, Mitchell, and Zhang}]{2020}
Wendi Ren, Yinghao Li, Hanting Su, David Kartchner, Cassie Mitchell, and Chao
  Zhang. 2020.
\newblock \href {https://doi.org/10.18653/v1/2020.findings-emnlp.334}
  {{Denoising Multi-Source Weak Supervision for Neural Text Classification}}.
\newblock \emph{Findings of the Association for Computational Linguistics:
  EMNLP 2020}.

\bibitem[{Rezende and Mohamed(2016)}]{rezende2016variational}
Danilo~Jimenez Rezende and Shakir Mohamed. 2016.
\newblock \href {http://arxiv.org/abs/1505.05770} {{Variational Inference with
  Normalizing Flows}}.

\bibitem[{Sedova et~al.(2021)Sedova, Stephan, Speranskaya, and
  Roth}]{sedova2021knodle}
Anastasiia Sedova, Andreas Stephan, Marina Speranskaya, and Benjamin Roth.
  2021.
\newblock \href {http://arxiv.org/abs/2104.11557} {{Knodle: Modular Weakly
  Supervised Learning with PyTorch}}.

\bibitem[{Takamatsu et~al.(2012)Takamatsu, Sato, and
  Nakagawa}]{takamatsu-etal-2012-reducing}
Shingo Takamatsu, Issei Sato, and Hiroshi Nakagawa. 2012.
\newblock \href {https://aclanthology.org/P12-1076} {{Reducing Wrong Labels in
  Distant Supervision for Relation Extraction}}.
\newblock In \emph{Proceedings of the 50th Annual Meeting of the Association
  for Computational Linguistics (Volume 1: Long Papers)}, pages 721--729, Jeju
  Island, Korea. Association for Computational Linguistics.

\bibitem[{Tran et~al.(2019)Tran, Vafa, Agrawal, Dinh, and
  Poole}]{tran2019discrete}
Dustin Tran, Keyon Vafa, Kumar~Krishna Agrawal, Laurent Dinh, and Ben Poole.
  2019.
\newblock \href {http://arxiv.org/abs/1905.10347} {{Discrete Flows: Invertible
  Generative Models of Discrete Data}}.

\bibitem[{Varma et~al.(2019)Varma, Sala, He, Ratner, and
  R{\'{e}}}]{varma2019learning}
Paroma Varma, Frederic Sala, Ann He, Alexander Ratner, and Christopher
  R{\'{e}}. 2019.
\newblock \href {http://arxiv.org/abs/1903.05844} {{Learning Dependency
  Structures for Weak Supervision Models}}.

\bibitem[{Zhang et~al.(2021)Zhang, Yu, Li, Wang, Yang, Yang, and
  Ratner}]{zhang2021wrench}
Jieyu Zhang, Yue Yu, Yinghao Li, Yujing Wang, Yaming Yang, Mao Yang, and
  Alexander Ratner. 2021.
\newblock \href {https://openreview.net/forum?id=Q9SKS5k8io} {{WRENCH}: A
  comprehensive benchmark for weak supervision}.
\newblock In \emph{Thirty-fifth Conference on Neural Information Processing
  Systems Datasets and Benchmarks Track}.

\bibitem[{Ziegler and Rush(2019)}]{ziegler2019latent}
Zachary~M Ziegler and Alexander~M Rush. 2019.
\newblock \href {http://arxiv.org/abs/1901.10548} {{Latent Normalizing Flows
  for Discrete Sequences}}.

\bibitem[{Zoph et~al.(2016)Zoph, Yuret, May, and Knight}]{zoph2016transfer}
Barret Zoph, Deniz Yuret, Jonathan May, and Kevin Knight. 2016.
\newblock \href {http://arxiv.org/abs/1604.02201} {{Transfer Learning for
  Low-Resource Neural Machine Translation}}.

\end{thebibliography}
\bibliographystyle{acl_natbib}

\appendix

\section{Additional Data Description}

\subsection{Preprocessing}
\label{subsec:app_preprocessing}

A few steps were performed, to create a unified data format. 
The crucial difference to other papers is that we removed duplicated samples.  There were two cases. 
Either there were very little duplicates or the duplication occurred because of the programmatic data generation, thus not resembling the real data generating process. 
Most notably, in the spouse data set $60\%$ of all data points are duplicates.
Furthermore, we only used rules which occurred more often than a certain threshold as it is impossible to learn densities on only a handful of examples. The threshold is 
In order to have unbiased baselines, we ran the baseline experiments on the full set of rules and the reduced set of rules and took the best performing number.

\subsection{IMDb rules}
\label{subsec:app_imdb_keywords}

The labeling functions for the IMDb dataset are defined by keywords. We manually chose the keywords.  We defined them in such a way that their meaning has rather little semantic overlap. The keywords are shown in table \ref{table:imdb_keywords}.

 \begin{table}
\centering
\begin{tabular}{cc}
\toprule
 positive & negative \\
\midrule
     beautiful &          poor \\
      pleasure & disappointing \\
recommendation &     senseless \\
      dazzling &   second-rate \\
   fascinating &         silly \\
     hilarious &        boring \\
    surprising &      tiresome \\
   interesting & uninteresting \\
   imaginative &      dreadful \\
      original &      outdated \\
\bottomrule
\end{tabular}
\caption{Keywords used to create rules for the IMDb dataset.}
\label{table:imdb_keywords}
\end{table}

\subsection{Labeling Function Correlations}
\label{sec:app_lf_correlation}
In order to use labeling functions for weakly supervised learning, it is important to know the correlation of labeling functions to i) derive methods to combine them and ii) help to understand phenomena of the model predictions.

Thus we decided to add correlation plots. More specifically, we use the Pearson Correlation coefficient.

\begin{figure*}
\centering
\begin{subfigure}{.48\textwidth}
  \centering
  \includegraphics[width=\linewidth]{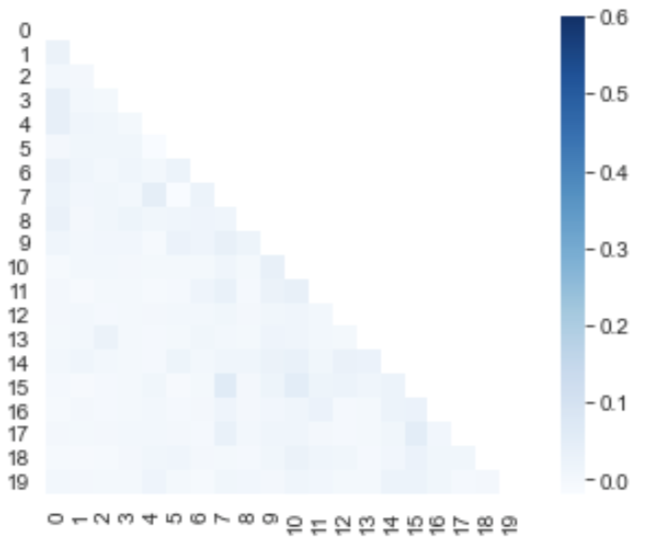}
  \caption{IMDb}
\end{subfigure}%
$ $ $ $ $ $ $ $
\begin{subfigure}{.48\textwidth}
  \centering
  \includegraphics[width=\linewidth]{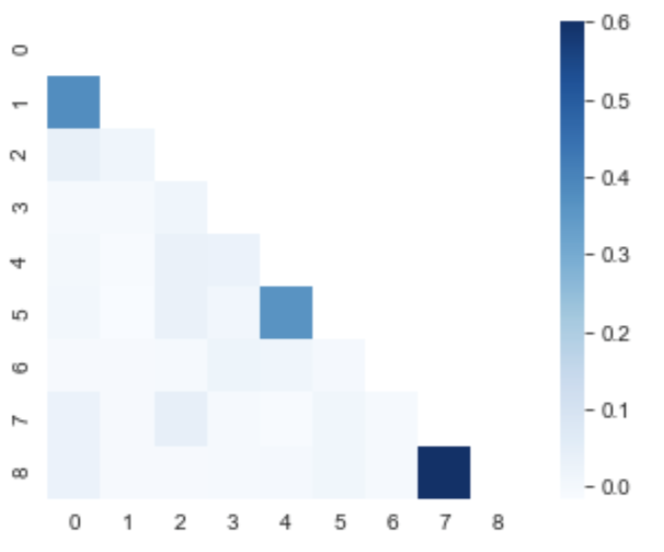}
  \caption{Spouse}
\end{subfigure}
\begin{subfigure}{.48\textwidth}
  \centering
  \includegraphics[width=\linewidth]{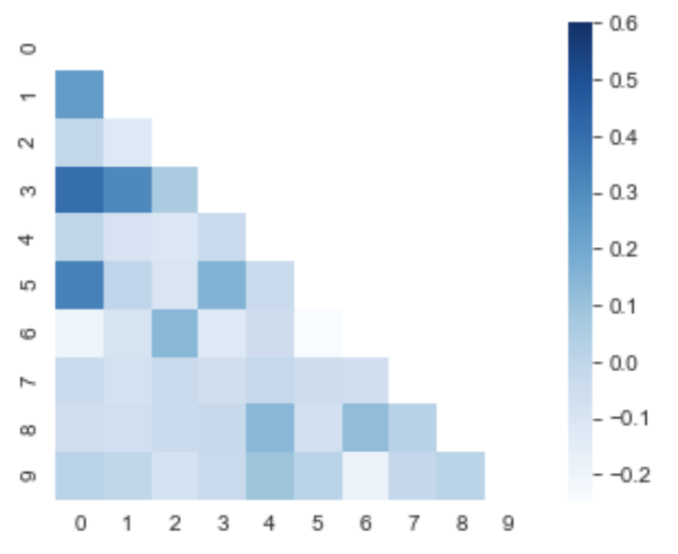}
  \caption{YouTube}
\end{subfigure}%
$ $ $ $ $ $ $ $
\begin{subfigure}{.48\textwidth}
  \centering
  \includegraphics[width=\linewidth]{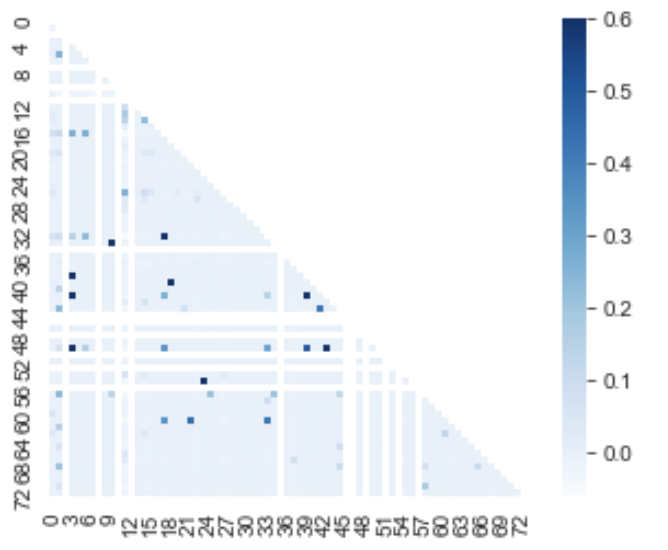}
  \caption{SMS}
\end{subfigure}
\begin{subfigure}{.48\textwidth}
  \centering
  \includegraphics[width=\linewidth]{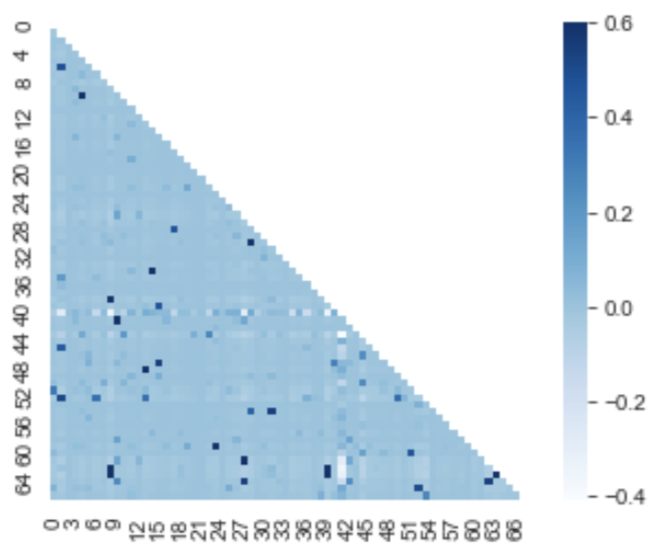}
  \caption{Trec}
\end{subfigure}
\label{fig:test}
\end{figure*}

\section{Additional Implementationial Details}
\subsection{Architecture}
\label{sec:app_detailed_architecture}

As mentioned in section \ref{sec:modelling}, the backbone of our flow is RealNVP architecture, which we introduced in section \ref{sec:related_work}. 
With sticking to the notation in formula \ref{eq:real_nvp} the network layers to approximate the functions $s$ and $t$ are shown below

\vspace{\baselineskip}
\small{
\begin{lstlisting}[language=python,
				 frame=leftline,keywordstyle=\color{blue},
				 commentstyle=\ttfamily\itshape\color{gray}, showstringspaces=false,
				 xleftmargin=2pt,
				 xrightmargin=1pt,
				 gobble=4,
				 frame=single, basicstyle=\scriptsize\ttfamily,
				 numbers=left]
s = nn.Sequential(
    nn.Linear(dim, hidden_dim),
    nn.LeakyReLU(),
    nn.BatchNorm1d(hidden_dim),
    nn.Dropout(0.3),
    nn.Linear(hidden_dim, dim),
    nn.Tanh()
)
t = nn.Sequential(
    nn.Linear(dim, hidden_dim),
    nn.LeakyReLU(),
    nn.BatchNorm1d(hidden_dim),
    nn.Dropout(0.3),
    nn.Linear(hidden_dim, dim),
    nn.Tanh()
)
\end{lstlisting}}
\vspace{\baselineskip}
\normalsize

Hyperparameters are the depth, i.e. number of stacked layers, and the hidden dimension.

\subsection{WeaNF-M Sampling}
\label{sec:app_sampling_details}

For the mixed model WeaNF-M the sampling process becomes rather complicated. 

Next up, the code to produce the convex combination $\alpha_1, \dots, \alpha_t$ is shown.
The input tensor takes values in $\{0, 1\}$ and has shape $b \times t$ where $b$ is the batch size and $t$ the number of labeling functions.Note that some mass is put on every labeling functions. We realized that this bias imrpoves performance.

\vspace{\baselineskip}
\small{
\begin{lstlisting}[language=python,
				 frame=leftline,keywordstyle=\color{blue},
				 commentstyle=\ttfamily\itshape\color{gray}, showstringspaces=false,
				 xleftmargin=2pt,
				 xrightmargin=1pt,
				 gobble=4,
				 frame=single, basicstyle=\scriptsize\ttfamily,
				 numbers=left]
def weight_batch(self, batch_y: torch.Tensor):
    """Returns weighting array forming convex sum. 
       Shape: (batch_dim, num_rules)
    """
    batch_y = batch_y.float()
    batch_y += 0.1 * torch.ones(batch_y.shape)
    batch_y = batch_y * torch.rand(batch_y.shape)
    row_sum = batch_y.sum(axis=1, keepdims=True)
    nbatch_y = batch_y / row_sum
    return nbatch_y
\end{lstlisting}}
\vspace{\baselineskip}
\normalsize




\end{document}